\documentclass[12pt]{article}

\PassOptionsToPackage{hidelinks,unicode,hypertexnames=false}{hyperref}

\usepackage[margin=1in]{geometry}
\usepackage[utf8]{inputenc}
\usepackage[T1]{fontenc}
\usepackage[authoryear]{natbib}
\usepackage{booktabs}
\usepackage{graphicx}
\usepackage{amsmath}
\usepackage{tabularx}
\usepackage{array}
\usepackage{longtable}
\usepackage{float}
\usepackage{nameref}
\usepackage{hyperref}
\usepackage{silence}
\newcolumntype{Y}{>{\raggedright\arraybackslash}X}

\hypersetup{
  colorlinks=true,
  linkcolor=black,
  citecolor=black,
  urlcolor=black
}

\newcommand{\seeref}[1]{\hyperref[#1]{\nameref{#1}}}

\title{Hybrid topic modelling for computational close reading: Mapping narrative themes in Puškin's \emph{Evgenij Onegin}}
\author{
Angelo Maria Sabatini\\
The BioRobotics Institute, Scuola Superiore Sant'Anna\\
Pisa, Italy\\
\texttt{angelo.sabatini@santannapisa.it}
}
\date{}

\begin{document}

\maketitle

\begin{abstract}
This study presents a hybrid topic modelling framework for computational literary analysis that integrates Latent Dirichlet Allocation (LDA) with sparse Partial Least Squares Discriminant Analysis (sPLS–DA) to model thematic structure and longitudinal dynamics in narrative poetry. As a case study, we analyse \emph{Evgenij Onegin}—Aleksandr S. Puškin’s novel in verse—using an Italian translation, testing whether unsupervised and supervised lexical structures converge in a small-corpus setting.

The poetic text is segmented into thirty-five documents of lemmatised content words, from which five stable and interpretable topics emerge. To address small-corpus instability, a multi-seed consensus protocol is adopted. Using sPLS–DA as a supervised probe enhances interpretability by identifying lexical markers that refine each theme. Narrative hubs—groups of contiguous stanzas marking key episodes—extend the bag-of-words approach to the narrative level, revealing how thematic mixtures align with the poem's emotional and structural arc.

Rather than replacing traditional literary interpretation, the proposed framework offers a computational form of close reading, illustrating how lightweight probabilistic models can yield reproducible thematic maps of complex poetic narratives, even when stylistic features such as metre, phonology, or native morphology are abstracted away. Despite relying on a single lemmatised translation, the approach provides a transparent methodological template applicable to other high-density literary texts in comparative studies.
\end{abstract}

\noindent\textbf{Keywords:} Computational Close Reading; Hybrid Topic Modelling; Latent Dirichlet Allocation; Literary Translation; Narrative Poetry; Sparse Partial Least Squares-Discriminant Analysis

\section*{Introduction}
\label{sec:intro}

Within computational literary studies, thematic structure is one of the central organising principles of literary narrative, yet its articulation across the unfolding of a text often remains implicit and difficult to formalise. Probabilistic topic modelling—and Latent Dirichlet Allocation (LDA) in particular—offers a powerful framework for the unsupervised exploration of large text collections \citep{BleiNgJordan2003}. In an LDA model, documents are represented as mixtures of latent topics, that is, probabilistic distributions over words inferred from their co-occurrence patterns. Each topic yields a ranked list of salient words that can often be interpreted as markers of an underlying theme.

While this approach has proven particularly effective for scientific and technical writing \citep{Blei2012}, its application to literary texts (especially poetry) poses additional challenges. In poetry, words co-occur in figurative, often ambiguous contexts; themes interlace through rhythm, sound, and voice, resisting the simplifying assumptions of the bag-of-words model. Yet several studies have shown how topic modelling can complement close reading by revealing recurrent semantic patterns that support interpretation \citep{JockersMimno2013,RoeGladstoneMorrissey2016,Schoech2017,NavarroColorado2018}.

Across large-scale literary corpora, topic modelling has been used to recover thematic patterns and their relation to metadata. \citet{JockersMimno2013}, analysing over 3,000 Anglophone novels (1750–1899), identified gendered uses of recurrent themes, combining permutation and bootstrap validation with a supervised probe to assess robustness. \citet{RoeGladstoneMorrissey2016} applied LDA to Diderot and d’Alembert’s \emph{Encyclopédie}, comparing topic-based “discursive formations” with the editors’ taxonomy, and highlighting the interpretive dependence on labelling choices and model size. Later studies extended this approach to drama and poetry. \citet{Schoech2017} explored 391 French plays (1630–1789), examining how segmentation and topic number affect interpretability. \citet{NavarroColorado2018}, analysing over 5,000 Spanish Golden Age sonnets, distinguished between “themes” and “motifs,” proposing both semantic and phonetic motifs while noting that interpretability ultimately depends on close reading. These examples illustrate both the promise and the limits of topic modelling for verse: figurative language, metre, and sound patterning strain the bag-of-words assumption and often yield opaque topic lists \citep{Rhody2012,Herbelot2015}. Hence, interpretive validation through close reading remains essential.

Recent computational approaches have broadened the analytical repertoire of literary studies beyond topic modelling \citep{Dieng2020}. Techniques such as sentiment and emotion analysis, named-entity recognition, and word-embedding models have been increasingly applied to prose and poetry, widening the spectrum of methods aimed at recovering affective or semantic structures from literary corpora \citep{Picca2024}. 

The present study deliberately retains a probabilistic topic model in order to preserve the interpretability of lexical themes within a single text. More closely aligned with the purpose of the present study, \citet{SaccentiTenori2012,RomanoConversano2025} apply unsupervised and supervised classification algorithms to poetic corpora, showing how lexical features can be predictive of thematic content even in a single-poem setting (\emph{Divina Commedia}). The present study therefore proposes a hybrid workflow that combines probabilistic topic modelling and supervised lexical discrimination to support computational close reading of a single poetic text.

While topic modelling provides scalable access to patterns invisible to close reading, the present study adopts an opposite vantage point. LDA is applied not because the text is unreadable at scale—the typical rationale for computational analysis—but precisely because it is fully legible. This shift—from large-scale thematic discovery to within-text narrative mapping—places topic modelling in the service of computational close reading rather than distant reading.

The goal is to test whether LDA remains informative and reliable within a single, well-defined work, when interpretability takes precedence over large-scale discovery. Specifically, we apply LDA to Aleksandr S. Puškin’s novel in verse, \emph{Evgenij Onegin}, to formalise reader-salient themes and make their within-text dynamics explicit. Since, in the case of \emph{Evgenij Onegin}, no predefined longitudinal labels exist beyond chapters, we turn to unsupervised modelling (LDA) to induce themes and their dynamics. 

For the present study we rely on the recent Italian translation by Giuseppe Ghini \citep{Pushkin2021}. Because the present analysis focuses on lexical thematic structure rather than on prosodic or phonological features, a high-quality literary translation provides a suitable basis for computational modelling. Ghini’s version preserves the stanzaic architecture of the poem and renders its lexical and stylistic register with high philological fidelity, while maintaining rhythmic coherence through unrhymed nine-syllable lines (\emph{novenari sciolti}).

To validate and enrich this interpretation, we propose a hybrid procedure in the spirit of \citet{SaccentiTenori2012}: a sparse Partial Least Squares Discriminant Analysis (sPLS–DA) is applied as a supervised probe to the same document–term matrix used by LDA . The method exploits the thematic labels inferred by LDA itself—not for primary classification, but to test whether the discriminant words identified by sPLS–DA confirm or refine LDA’s topics. This combination treats LDA and sPLS–DA as complementary components of a single interpretive framework. LDA provides probabilistic coherence, while sPLS–DA introduces discriminative specificity. Together they enable a computational form of close reading that bridges algorithmic inference and literary interpretation within a transparent and reproducible workflow.

\subsection*{\textit{Evgenij Onegin}}

More than a literary masterpiece, \textit{Evgenij Onegin} stands as a cornerstone of modern Russian literary language \citep{Carpi2020}. Written between 1823 and 1831, it comprises eight canonical parts (chapters). Its formal invention—the Onegin stanza, a 14-line iambic tetrameter with the characteristic rhyme scheme $\mathrm{aBaBccDDeFFeGG}$ and alternating feminine and masculine endings—provides a highly regular metrical frame for what Puškin himself called a "novel in verse." Occasionally, numbered stanzas are intentionally missing, so the traditional count of 389 stanzas corresponds to 366 with textual content. Within this formal constraint Puškin achieves striking stylistic variety and rapid shifts of register. The tonal palette is enriched by multilingual insertions—mostly French, but also other European languages—woven into dialogues, letters, and narrator digressions, producing a heteroglossic texture in which multiple registers and voices intertwine. 

\paragraph{Plot of the verse novel}
The story opens with the presentation of Evgenij Onegin, a young aristocrat from Saint Petersburg—a friend of the narrator, worldly yet already bored by pleasures that come too easily. Summoned to the countryside by his uncle’s death, he inherits an estate and meets Vladimir Lenskij, a Romantic poet recently returned from Germany, who introduces him to the neighbouring Larin family: the lively, down-to-earth Ol’ga (Lenskij’s fiancée) and her elder sister Tat’jana, quiet, thoughtful, and fond of books.

Tat’jana falls in love with Evgenij at first sight and, in a key night scene, writes him an openhearted letter. Evgenij’s reply—delivered during a conversation in the garden—refuses her love and warns her to be cautious: he claims to be unfit for constancy and tells her that her honesty would only bring her pain. Soon after, during Tat’jana’s name-day celebration, things turn awry: irritated by provincial gossip, Evgenij deliberately flirts with Ol’ga. Honour codes do the rest. Despite his regrets the next morning, the duel takes place, and Evgenij kills his friend.

Overcome by guilt and restlessness, Evgenij leaves the district. Some time later, Tat’jana, now alone in her parents’ house after her sister’s quick marriage, wanders through Evgenij’s empty estate, reading his marginalia and notebooks; she finds not a coherent self but a patchwork of borrowed poses. Time passes. After leaving the countryside for Moscow, Tat’jana wins the affection of a general of noble rank, who marries her. Years later, Evgenij returns to Saint Petersburg and recognises her at a ball—now transformed into a poised and distant lady whose composure turns his former indifference into overwhelming desire.

In the final part, Evgenij becomes the pursuer, sending her a letter and seeking a private meeting; the roles are reversed. When they finally meet, Tat’jana admits she still loves him but refuses to betray her vows. The verse novel closes on this moral decision rather than reconciliation, leaving Evgenij's fate unresolved.

To relate the computational results to Puškin’s narrative design, we introduce the notion of \emph{narrative hubs}. Narrative hubs are contiguous stanza groups corresponding to salient episodes commonly recognised by readers and literary criticism. This allows us to test whether the latent themes extracted computationally align with the verse novel’s narrative structure at pivotal moments.

Thirteen illustrative hubs are listed in Table~\ref{tab:EO-hubs}; they serve as fixed reference points for visual summaries and are revisited in \seeref{sec:results}, where four of them are discussed in detail.


\begin{table}[ht]
\centering
\caption{Selected list of narrative hubs in \textit{Evgenij Onegin}, with corresponding chapter (ch.) and stanza ranges.}
\label{tab:EO-hubs}
\begin{tabularx}{\linewidth}{@{}rYll@{}}
\toprule
\textbf{ID} & \textbf{Hub label} & \textbf{Start (ch., stanza)} & \textbf{End (ch., stanza)} \\
\midrule
$H_1$ & A typical day in Evgenij's life & (1,~XV) & (1,~XXVIII) \\
$H_2$ & The friendship between Evgenij and the Poet & (1,~XLV) & (1,~L) \\
$H_3$ & Tat'jana's introduction & (2,~XXIV) & (2,~XXIX) \\
$H_4$ & Tat'jana falls in love & (3,~VII) & (3,~XXI) \\
$H_5$ & Tat’jana’s letter & (3,~XXXI) & (3,~XXXI) \\
$H_6$ & Evgenij's reply & (4,~XI) & (4,~XVII) \\
$H_7$ & Tat’jana’s dream & (5,~XI) & (5,~XXIV) \\
$H_8$ & Fest at Larins' home & (5,~XXIX) & (5,~XLV) \\
$H_9$ & The duel and Lenskij’s death & (6,~XXVI) & (6,~XXXV) \\
$H_{10}$ & In memory of Lenskij & (6,~XXXVI) & (6,~XLVI) \\
$H_{11}$ & Journey to Moscow & (7,~XXVIII) & (7,~XXXV) \\
$H_{12}$ & Evgenij’s letter & (8,~XXXII) & (8,~XXXII) \\
$H_{13}$ & Tat’jana’s reply & (8,~XLII) & (8,~XLVII) \\
\bottomrule
\end{tabularx}
\end{table}

The paper is structured as follows. Section~\seeref{sec:methods} describes the corpus, stanza-based segmentation, and pre-processing pipeline. Section~\ref{sec:results} presents the induced topics, thematic map, and longitudinal dynamics, including sensitivity analyses and supervised probes (sPLS–DA). Section~\ref{sec:discussion} discusses interpretability, relates findings to prior work, and outlines limitations. Finally, Section~\seeref{sec:conclusion} summarises key results and suggests directions for future comparative work.

\section*{Methods}
\label{sec:methods}

\subsection*{Corpus preparation and preprocessing}
\label{sec:methods-corpus}

The corpus consists of the Italian translation of \textit{Evgenij Onegin} by Giuseppe Ghini \citep{Pushkin2021}. The printed text of the eight canonical chapters was scanned and processed with the open-source tool \texttt{ocrmypdf}, followed by a manual revision to correct minor OCR errors. All preprocessing and analyses were performed in R (version~4.4.1). Each stanza was assigned a unique (chapter, stanza) identifier, while internal line segmentation was preserved to allow optional per-line analyses. Dedication and chapter-opening epigraphs were annotated separately and excluded from quantitative modelling. Formal features of Puškin’s verse were also annotated, including collapsed stanzas (distinct identifiers sharing the same lines) and empty stanzas (those without any verse or consisting solely of dots).

Two specific traits of Ghini’s translation deserve mention. First, several proper names—and a smaller set of common nouns—appear in transliterated form, with an apostrophe marking the soft sign of the Cyrillic alphabet. Second, the translation contains numerous foreign borrowings, mainly from French; these were treated as ordinary tokens throughout the analysis.

Tokenisation, part-of-speech (PoS) tagging, and lemmatisation were carried out with UDPipe (model: \texttt{italian-isdt-ud-2.5-191206}) \citep{udpipe}. Before tagging, a small set of hyphenated compounds (e.g., \textit{comme-il-faut}, \textit{bon-ton}) was split to improve tokenisation. Capitalised tokens with internal apostrophes were manually inspected to distinguish Italian elisions from transliterations of the soft sign (e.g., \textit{Tat'jana}, \textit{Ol'ga}). For such tokens, the apostrophe was temporarily removed before tagging to facilitate lemmatisation, then restored via deterministic mapping; proper nouns (PROPN) were verified for consistent labelling.

A four-step curation was applied to UDPipe outputs. First, targeted PoS fixes corrected recurrent mislabellings among French, Latin, English, and occasional Russian tokens (e.g., \textit{vous}, \textit{primis}, \textit{poor}, \textit{domovoj}). Second, capitalisation was checked and restored for entities such as \textit{Dio} and \textit{Musa}. Third, for lemmas ending in \textit{-are/-ere/-ire} that UDPipe often misclassified, we reassigned them as VERB under an evidence-based rule: if the proportion of VERB tags for that lemma was $\ge 0.60$, there were at least two occurrences tagged as VERB, and lemma length was $\ge 5$ characters, the VERB tag was enforced for all instances; otherwise, the original tags were retained. Fourth, noisy lemmas were detected and corrected together with their tag (e.g., \textit{tristere}$\rightarrow$\textit{triste}/ADJ, \textit{epigrammo}$\rightarrow$\textit{epigramma}/NOUN, \textit{njanjo}$\rightarrow$\textit{njanja}/NOUN).

From the PoS-tagged corpus, four categories of content words were initially considered as candidates for the final vocabulary: NOUN, PROPN, ADJ, and VERB. A frequency threshold removed all lemmas occurring fewer than three times. Proper names preserved capitalisation, whereas nouns, adjectives, and verbs were lowercased. Manual inspection of all lemmas occurring three or more times confirmed the absence of noisy forms or PoS misclassifications and the consistent capitalisation of proper nouns.

From the PoS-tagged corpus, four categories of content words were initially considered as candidates for the final vocabulary: NOUN, PROPN, ADJ, and VERB. A frequency threshold removed all lemmas occurring fewer than three times. Extremely low-frequency terms (hapax and dislegomena) were excluded in order to stabilise the estimation of topic–term distributions, a common preprocessing step in topic modelling pipelines \citep{Silge 2017}. In narrative texts, such items frequently correspond to minor character names or incidental references and therefore contribute little to stable thematic structure. Manual inspection of all lemmas occurring three or more times confirmed the absence of noisy forms or PoS misclassifications and the consistent capitalisation of proper nouns. Proper names preserved capitalisation, whereas nouns, adjectives, and verbs were lowercased. 

The inclusion of VERB lemmas was explicitly evaluated during preprocessing. Preliminary exploratory runs and inspection of high-frequency verbal lemmas indicated that their inclusion tended to foreground narrative progression rather than stabilise descriptive thematic fields. Since the primary goal of the present study was to formalise reader-salient thematic content, we ultimately excluded VERB lemmas and retained only NOUN, PROPN, and ADJ lemmas in the final vocabulary. This choice privileges thematic interpretability over full lexical coverage, and should be understood as a modelling decision rather than as a claim about the intrinsic literary salience of verbal forms. Stop-word removal was limited to a short manually curated list of frequent function words whose semantic contribution was negligible for thematic interpretation (e.g., \emph{altro} ‘other’, \emph{grande} ‘big’, \emph{cosa} ‘thing’). This conservative filtering was adopted to avoid removing stylistically meaningful lexical material.

Within the metrically coherent structure of the verse novel, the 14-line Onegin stanza represents its smallest structural unit. For computational purposes, however, individual stanzas are too short to yield a stable lexical signal. We therefore adopted an aggregation strategy that groups contiguous stanzas into larger units. Specifically, the verse novel was segmented into contiguous, ordered blocks such that each of the eight canonical chapters decomposes into an integer number of similarly sized blocks. This procedure resulted in a total of 35 blocks (documents), each comprising ten or eleven stanzas. Such segmentation provides documents with an adequate lexical signal for modelling within a standard LDA framework, thereby avoiding the complexities associated with specialised short-text models or word embeddings (cf.~\citet{NavarroColorado2018}). Block–lemma counts were aggregated into a Document–Term Matrix (DTM). The DTM was stored in long format with the following fields: \texttt{block\_id}, encoded as \texttt{C\#\_B\#} (Chapter \#, Block \#); \texttt{lemma}; and raw term counts (\texttt{n}). The complete DTM is provided in reproducible form.

\subsection*{Topic modelling (LDA)}
\label{sec:methods-lda}

Latent Dirichlet Allocation (LDA) was applied to the DTM using the collapsed Gibbs sampler implemented in the R package \texttt{topicmodels} \citep{Grun2011}. We first conducted a light exploratory sweep over the number of topics $K$, in the spirit of \citet{GriffithsSteyvers2004}, inspecting both model fit (final log-likelihood) and the interpretability of top terms. Several coherence-based diagnostics were also explored; however, given the relatively small size of the corpus, these measures proved difficult to interpret and did not provide stable guidance on the optimal value of $K$. Privileging stability and semantic clarity, we therefore fixed $K=5$ for all subsequent analyses. This value yielded semantically coherent topics while maintaining sufficient thematic differentiation across the corpus, as further supported by the effective number of topics $N_{\text{eff}}$ reported below. Dirichlet symmetric priors were set to $\alpha=2$ (document–topic) and $\beta=0.15$ (topic–term), and kept fixed during sampling. With $K=5$, $\alpha=2$ yields a Dirichlet concentration ($K\alpha=10$) that encourages mixed (non-spiky) $\gamma$ distributions on short documents. This value of $\alpha$ was chosen deliberately to counteract the short length of the blocks, ensuring that all topics are represented to some extent and preventing the emergence of spurious or outlier topics. Simultaneously, $\beta=0.15$ provides a high degree of smoothing over the reduced vocabulary, ensuring that rare or idiosyncratic lemmas do not dominate a topic; neither prior was submitted to hyperparameter optimisation.

Because the Gibbs sampler is stochastic and topic indices are non-identifiable up to permutation (“label switching”), different random initialisations can converge to distinct local optima and arbitrary labelling. We therefore adopted a multi-run ensemble with post-hoc topic alignment and consensus estimation, following stability-oriented practice in the literature \citep{Greene2014,Chuang2015}. We ran 100 independent chains (each 4{,}000 iterations; burn-in 2{,}000; thinning 100). For every run we extracted the matrices $\beta$ (term–topic weights) and $\gamma$ (document–topic mixtures). To reduce sensitivity to initial settings and make runs comparable, we identified as reference the run that maximised the final log-likelihood; all other runs were aligned to the reference by solving a Hungarian assignment \citep{Kuhn1955,Munkres1957} on a composite similarity:
$$
S = w_\beta S_\beta + w_\gamma R_\gamma,
$$
where $S_\beta$ is the Jaccard overlap of top–30 lemma sets (Jaccard@30) and $R_\gamma$ is the Spearman rank correlation between columns of $\gamma$. Equal weighting ($w_\beta=w_\gamma=0.5$) was adopted to prevent either lexical content ($S_\beta$) or document structure ($R_\gamma$) from unilaterally dominating the alignment, thereby balancing consistency checks.

Alignment quality was required to satisfy three criteria: mean Jaccard@30 $\ge 0.30$, mean Spearman $\ge 0.60$, and at least $0.60$ agreement between the $\beta$-only and $\gamma$-only mappings. Runs failing any criterion were excluded. For the retained runs, $\gamma$ matrices were column-realigned and averaged to obtain a consensus document–topic distribution. For each block we report the dominant and runner-up topics with their consensus $\gamma$ values. The top–30 terms per topic were taken from the reference run.

To summarise topic balance and provide an approximate score of the extent to which the extracted topics are well represented in the corpus, we computed the effective number of topics $N_{\text{eff}}$, equivalent to the inverse Simpson index \citep{Simpson1949}:
$$
N_{\text{eff}} = \dfrac{1}{\sum_{k=1}^K p_k^2},
$$
where $p_k$ is the proportion of documents whose dominant topic is the $k$-th topic. When $N_{\text{eff}}$ is reasonably close to $K$, all topics are present in the mixture; conversely, $N_{\text{eff}} < K$ indicates that a reduced number of topics dominate the corpus, suggesting that $K$ should be lowered. Although several coherence measures have been proposed in the literature (cf.~\citet{Roder2015}), $N_{\text{eff}}$ serves effectively as a stability diagnostic in our study.

\subsection*{Supervised validation (sPLS–DA)}
\label{sec:methods-spls}

We complemented the unsupervised analysis with a supervised probe to assess the internal consistency of LDA-derived topics. Following sPLS–DA practice for high-dimensional text and omics settings \citep{LeCao2011,RuizPerez2020}, we asked whether the dominant LDA topic assigned to each block can be recovered as a supervised class, and which lexical features are most discriminant under explicit supervision. sPLS–DA was used here as a validation probe rather than as a primary classification tool. This involved treating the thematic structure induced by LDA as the ground truth for classification: labels were taken from the LDA output and assigned to the DTM. Models were fitted with the R package \texttt{mixOmics} \citep{Rohart2017}.

To reduce multiclass complexity and focus on class-wise separability, we adopted a one-versus-rest (OvR) design, training five binary sPLS–DA classifiers (one per topic) and collapsing all other topics into the \texttt{other} class. Each OvR model used $n_{\mathrm{comp}}=2$ latent components and retained $\mathrm{keepX}=30$ variables per component (moderate sparsity to stabilise selection in a small corpus). Evaluation employed stratified 5-fold cross-validation repeated five times; we report overall and class-specific accuracy, as well as balanced accuracy, compared against a random baseline preserving class prevalence. This OvR step is purely diagnostic and mitigates the instability typical of direct multiclass fits in low-$n$ (35 documents) / high-$p$ (vocabulary size) regimes.

For lexicon consolidation, a single multiclass sPLS–DA was then refitted on the full dataset (using the same $n_{\mathrm{comp}}$ and $\mathrm{keepX}$ values adopted for the OvR diagnostic). Terms were ranked by (i) selection frequency across cross-validation folds and (ii) average absolute loading. In sPLS–DA, loadings are the weights defining each latent component as a linear combination of variables; larger absolute values indicate a stronger contribution of a term to discrimination.

To classify terms as \emph{core exclusive} vs. \emph{shared}, we defined for each term $t$ and topic $k$ the exclusivity index
$$
E(t, k) = \dfrac{f_{t, k}}{\sum_{k'} f_{t, k'}},
$$
where $f_{t, k}$ is the total count of $t$ in blocks whose dominant topic is $k$ (the denominator summing over all topics). We also computed a loading magnitude $\overline{|L|}(t)$ as the mean absolute loading of $t$ across the two latent components, averaged over folds where $t$ was selected, and a weighted exclusivity score
$$
W(t, k) = E(t, k)\times\overline{|L|}(t).
$$
Terms with $E(t, k)\ge\lambda$ were marked as \emph{core exclusive}; among these, we ranked by $W(t,k)$ and retained up to 20 terms per topic for the final dictionary used in hub reports and Supplementary Tables. Terms with $E(t, k)<\lambda$ were considered \emph{shared}. The threshold $\lambda=0.6$ was empirically determined to ensure lexical specificity while maintaining thematic coverage across the corpus.

Finally, we compared the sPLS–DA discriminant lists with the LDA top–30 terms by class-level Jaccard overlaps and visualised cross-method agreement in the heatmap LDA$\rightarrow$sPLS–DA. LDA captures latent co-occurrence structure, whereas sPLS–DA emphasises discriminant features under supervision; their intersection provides convergent evidence on the lexical structure of the verse novel.

\subsection*{Narrative hubs}
\label{sec:methods-hubs}

In addition to documents, we defined and used \emph{narrative hubs}, namely contiguous groups of stanzas chosen to match relevant episodes in the verse novel. Hubs did not need to coincide with the block segmentation used for LDA and could cut across block boundaries. To retain interpretability, hubs were selected so as to be neither too short nor excessively long (the set used here is introduced in Table~\ref{tab:EO-hubs}).

Each hub was summarised by a structured \emph{hub card} integrating a quantitative and a lexical layer, without introducing any new modelling. Let $H$ index a hub and $B(H)$ the set of blocks contributing at least one stanza to $H$. With weights
$$
w_{H,d} = \frac{\#\text{ stanzas from block } d \text{ in } H}{\#\text{ stanzas in } H}\qquad\Big(\sum_{d\in B(H)} w_{H,d}=1\Big),
$$
the hub-level document–topic mixture was defined as the weighted average
$$
\bar{\gamma}(H) = \sum_{d\in B(H)} w_{H,d}\gamma_d.
$$
We reported the dominant and runner-up components of $\bar{\gamma}(H)$.

Together with $\bar{\gamma}(H)$, we computed a $\beta$-based topic profile for each hub. Let $V$ denote the union of the per-topic top-term lists from LDA, and let $V_H=\{w\in V:c_{H,w}>0\}$, where $c_{H,w}$ is the frequency of lemma $w$ in hub $H$, be the subset occurring in $H$. With $\beta_{k,w}$ the term–topic weight, the (count-weighted) lexical mass per topic was
$$
S^{(\beta)}_{H,k} = \sum_{w \in V_H} c_{H,w}\beta_{k,w},
\qquad 
P^{(\beta)}_{H,k} = \frac{S^{(\beta)}_{H,k}}{\sum_{k'} S^{(\beta)}_{H,k'}} .
$$
We then defined
$$
P^{(\beta)}_{H} = \arg\max_k P^{(\beta)}_{H,k}
$$
as the $\beta$-dominant topic for hub $H$.

While LDA provided ranked term lists ($\beta$), such lists can be semantically heterogeneous. For interpretability, we grouped terms into compact, recurrent lexical clusters called \emph{subthemes}. These were human-curated, guided by close reading of the Italian text and supported by frequency patterns in the LDA dictionaries as well as by sPLS–DA discriminant lists (\seeref{sec:methods-spls}). For each hub we tallied counts derived from both LDA top terms and sPLS–DA core-exclusive items. A small "bucket" category was also included for generic lemmas whose presence contributes to the statistical signal but does not warrant assignment to a specific subtheme. This bucket is not a post-hoc stop-word list: its lemmas were informative for both LDA and sPLS–DA, yet they are grouped together to maintain clarity in the thematic taxonomy.

Hub cards thus acted as an interpretive interface between quantitative topic-model outputs and the verse novel’s narrative structure. Additional methodological details, including full per-theme dictionaries and the subtheme taxonomy, are provided in Supplementary Material~A. A complete example of a hub card illustrating theme composition and subtheme weighting is given in Supplementary Material~B.

\section*{Results}
\label{sec:results}

\subsection*{Computational framework}
\label{sec:results-framework}

Our segmentation strategy provided the thirty-five documents of the verse novel with an adequate lexical signal for robust LDA estimation. After lemmatisation and PoS filtering, documents contained a median of $165$ content words and $119$ distinct lemmas (min.~$94$; max.~$168$), providing sufficient co-occurrence signal while preserving within-text dynamics. The resulting Document–Term Matrix (DTM) was $35\times756$ (documents~$\times$~lemmas) with a sparsity of $s\approx84\%$. For comparison, an alternative segmentation into five–six-stanza blocks yielded a $70\times756$ matrix with $s\approx91\%$. The exclusion of the VERB category was empirically supported: its inclusion produced a DTM of $35\times1{,}018$ ($s\approx83\%$), which in exploratory runs yielded topics dominated by high-frequency auxiliary and narrative verbs, e.g., \emph{essere} ‘to be’, \emph{avere} ‘to have’, \emph{andare} ‘to go’, \emph{dire} ‘to say’. This confirmed the difficulty of separating thematic elements from truly descriptive content.

The reference model with $K=5$ consistently produced five interpretable topics, each with a distinct lexical profile. Across the retained runs (post–alignment thresholding), model stability was high: the mean Jaccard@30 per topic was $0.41$ (median~$0.41$, range~$0.32$–$0.49$), and the mean Spearman correlation between $\gamma$ columns was $0.91$ (median~$0.91$, range~$0.82$–$0.94$). Crucially, in every retained run, $100\%$ of topics received concordant assignments from the $\beta$-only and $\gamma$-only mappings. Consensus $\gamma$ distributions were then averaged over retained runs, yielding stable dominant and runner-up topics for each document. The mean effective number of topics was $N_{\text{eff}}=4.52$ (median~$4.59$, range~$3.45$–$4.92$), indicating strong topic balance.

Sensitivity analyses at $K=3$ and $K=8$ confirmed the expected extremes. At $K=3$, consensus was $100\%$, Jaccard@30 was $0.44$, and Spearman($\gamma$) was $0.89$, but the resulting $N_{\text{eff}}=2.70$ suggested thematic under-resolution. Conversely, at $K=8$, consensus dropped to $14\%$, Jaccard@30 to $0.28$, and Spearman($\gamma$) to $0.80$, with $N_{\text{eff}}=6.48$, indicating over-fragmentation and instability. For comparison, running the standard $K=5$ model with the inclusion of the VERB category preserved high stability (consensus~$100\%$, Jaccard@30~$0.40$, Spearman($\gamma$)~$0.92$) but resulted in a lower balance ($N_{\text{eff}}=4.13$), further supporting the exclusion of verbs for thematic clarity.

\begin{figure}[ht]
\centering
\includegraphics[width=0.9\linewidth]{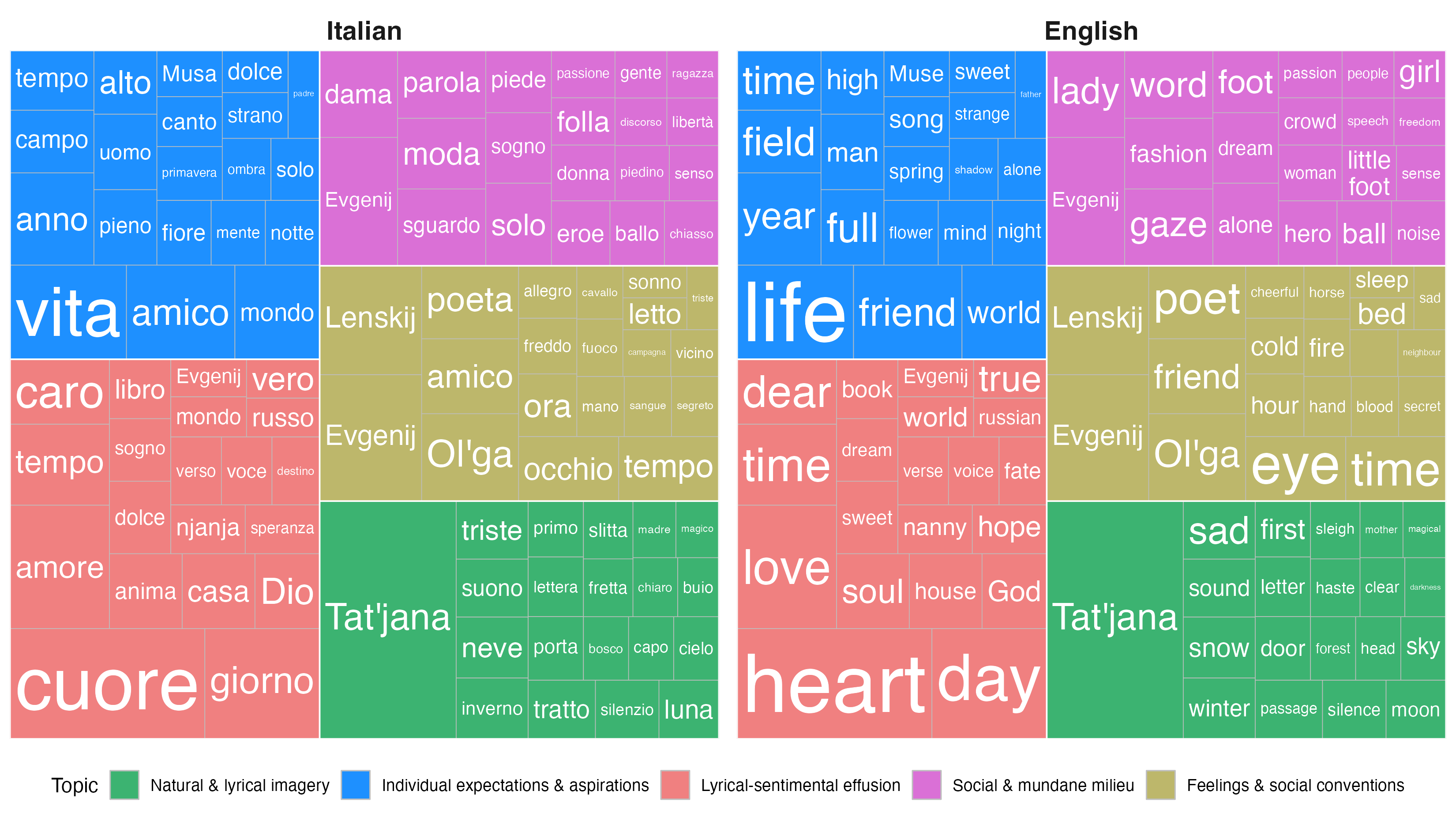}
\caption{Top~20 terms per topic (prevalence-weighted). Panel area per topic is proportional to its mean prevalence; tile area per lemma is proportional to its within-topic weight. Italian lemmas (left); English glosses (right).}
\label{fig:treemap}
\end{figure}

Table~\ref{tab:topics} lists the five induced topics ($\textsc{T}_1$–$\textsc{T}_5$) with their interpretive labels (themes), the number $n_k$ of documents where $\textsc{T}_k$ is dominant, and the mean prevalence $\bar{\gamma}_k$. To compactly present each theme’s lexicon, Fig.~\ref{fig:treemap} shows the \emph{top~20} lemmas per theme. Italian lemmas (IT) appear on the left; English (EN) glosses are provided on the right for expository purposes. Each theme was further subdivided into descriptive subthemes. The chosen subthemes and full per-theme dictionaries (top~30, IT+EN) are provided in Supplementary Material~A (Table~S1 and Tables~S2–S6, respectively).

\begin{table}[ht]
\centering
\caption{Topics with their interpretive label, per-topic dominant-document count $n_k$, and mean prevalence $\bar{\gamma}_k$.}
\label{tab:topics}
\begin{tabular}{@{}llrc@{}}
\toprule
\textbf{Topic} & \textbf{Label} & $\boldsymbol{n_k}$ & $\boldsymbol{\bar{\gamma}_k}$\\
\midrule
\textsc{T}$_1$ & Natural \& lyrical imagery & 7 & 0.194 \\
\textsc{T}$_2$ & Individual expectations \& aspirations & 8 & 0.196 \\
\textsc{T}$_3$ & Lyrical–sentimental effusion & 10 & 0.241 \\
\textsc{T}$_4$ & Social \& mundane milieu & 5 & 0.176 \\
\textsc{T}$_5$ & Feelings \& social conventions & 5 & 0.193 \\
\bottomrule
\end{tabular}
\end{table}

\begin{figure}[ht]
\centering
\includegraphics[width=1.1\linewidth]{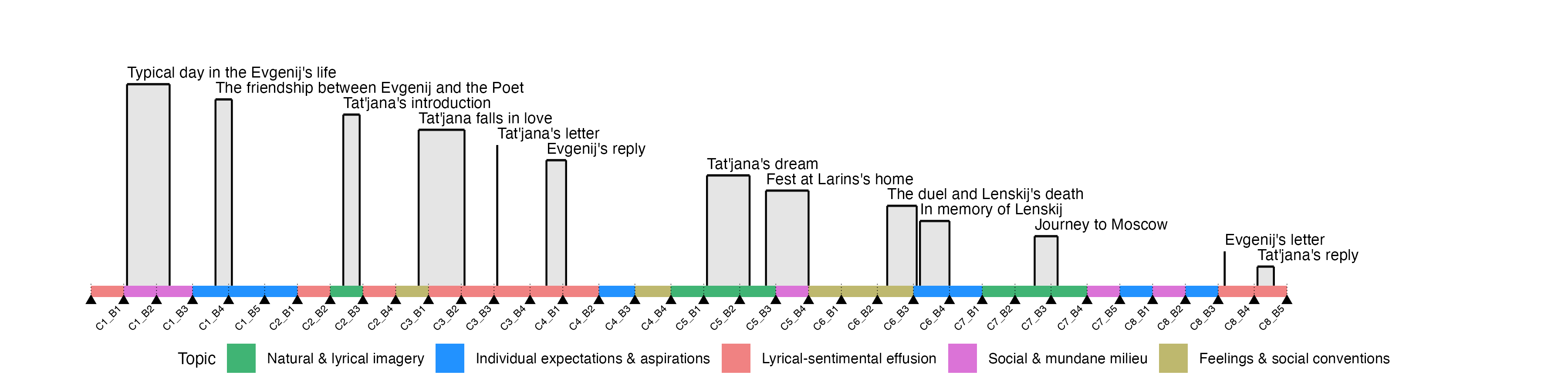}
\caption{Narrative map across documents. Grey bands mark the narrative hubs described in Table~\ref{tab:EO-hubs}.}
\label{fig:narrative-map}
\end{figure}

Figure~\ref{fig:narrative-map} maps the consensus topic mixture $\gamma_d$ across the documents of the corpus.

sPLS–DA OvR classifiers' performance is summarised in Table~\ref{tab:splsda-class}, where it is compared with class-prevalence–matched random baselines. Despite the small sample ($n=35$) and uneven class sizes (cf.~$n_k$ in Table~\ref{tab:topics}), the overall accuracy was $85\%$, and the overall balanced accuracy was $70\%$, with no class collapse (for reference, the random baseline yielded $70\%$ and $51\%$, respectively). Furthermore, the per-class balanced accuracy ($\text{bacc}$) substantially exceeded the random baseline ($\text{bacc}_0$) for all topics, confirming robust classification performance even in the presence of class imbalance \citep[e.g.][]{Brodersen2010}.

\begin{table}[ht]
\centering
\caption{Cross-validated performance of sPLS–DA OvR classifiers by topic.
Columns: mean accuracy ($\text{acc}$), balanced accuracy ($\text{bacc}$), and stratified random baselines ($\text{acc}_0$, $\text{bacc}_0$).}
\label{tab:splsda-class}
\begin{tabular}{lcccc}
\toprule
Topic & $\text{acc}$ & $\text{bacc}$ & $\text{acc}_0$ & $\text{bacc}_0$ \\
\midrule
Natural \& lyrical imagery  & 0.88 & 0.81 & 0.72 & 0.55 \\
Individual expectations \& aspirations  & 0.76 & 0.60 & 0.69 & 0.53 \\
Lyrical–sentimental effusion & 0.86 & 0.81 & 0.58 & 0.47 \\
Social \& mundane milieu & 0.89 & 0.68 & 0.79 & 0.55 \\
Feelings \& social conventions & 0.86 & 0.58 & 0.76 & 0.48 \\
\bottomrule
\end{tabular}
\end{table}

\begin{figure}[ht]
\centering
\includegraphics[width=0.6\linewidth]{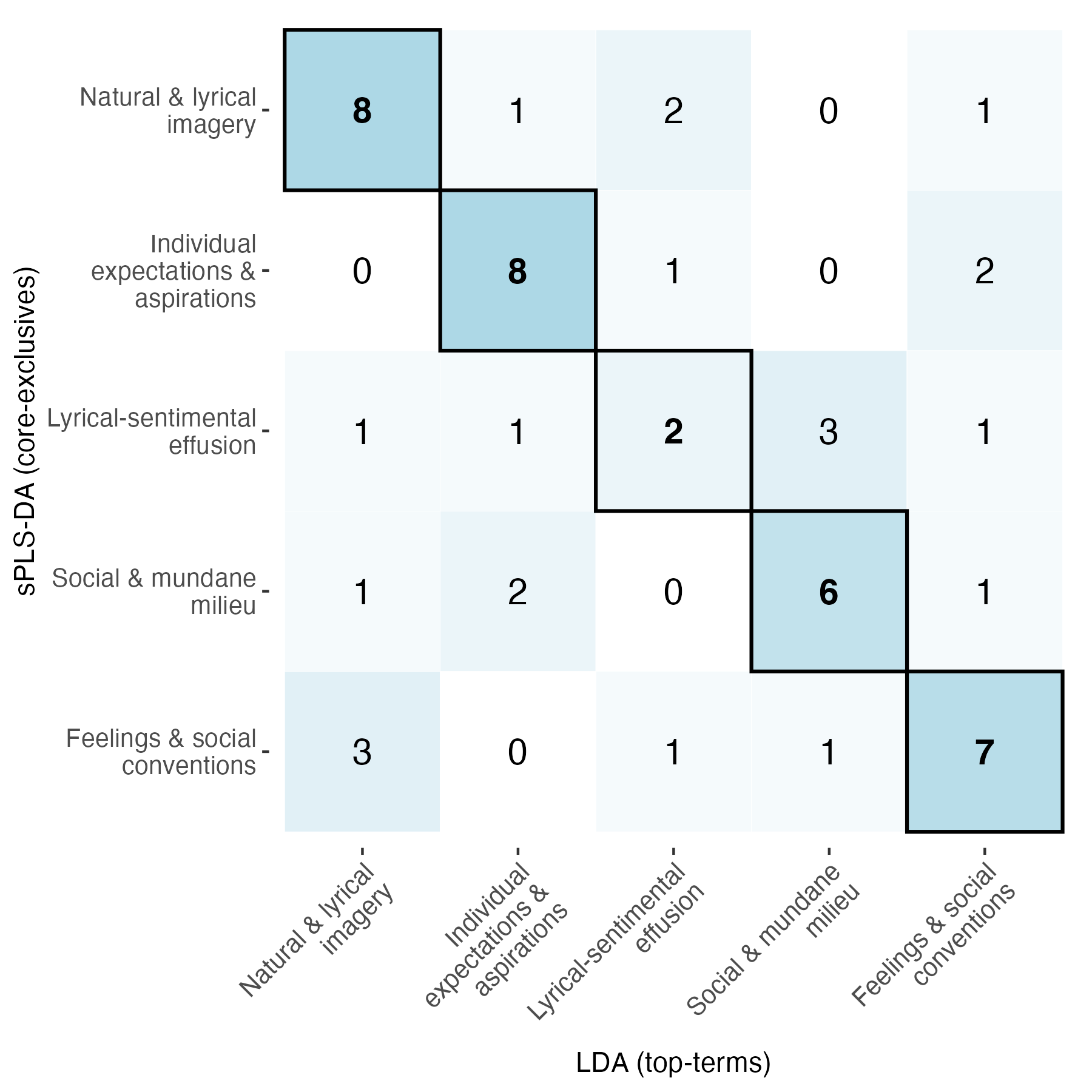}
\caption{Cross-over heatmap between sPLS–DA core-exclusive lemmas and LDA top lemmas. Diagonal values indicate within-theme overlap; off-diagonal values indicate cross-theme overlap.}
\label{fig:cross-over}
\end{figure}

Figure~\ref{fig:cross-over} illustrates how sPLS–DA core-exclusive lemmas distribute across LDA topics. Lemmas located along the diagonal belong to both models’ dictionaries for the same theme and are thus considered specific; those occurring in off-diagonal cells appear among the top lemmas of another topic and are classified as porous, i.e., items bridging adjacent themes. The complete list of these lemmas per theme is provided in Supplementary Material~A (Tables~S2–S6), together with their attribution to corresponding subthemes in Table~S1.

For illustration, the most porous class was \emph{Lyrical–sentimental effusion}, whose core-exclusive lemmas partly overlapped with those of \emph{Natural \& lyrical imagery} (\emph{cielo} ‘sky’), \emph{Individual expectations \& aspirations} (\emph{acqua} ‘water’), \emph{Social \& mundane milieu} (\emph{moglie} ‘wife’, \emph{gioia} ‘joy’, \emph{passione} ‘passion’), and \emph{Feelings \& social conventions} (\emph{fuoco} ‘fire’).

\begin{figure}[h]
\centering
\includegraphics[width=1\linewidth]{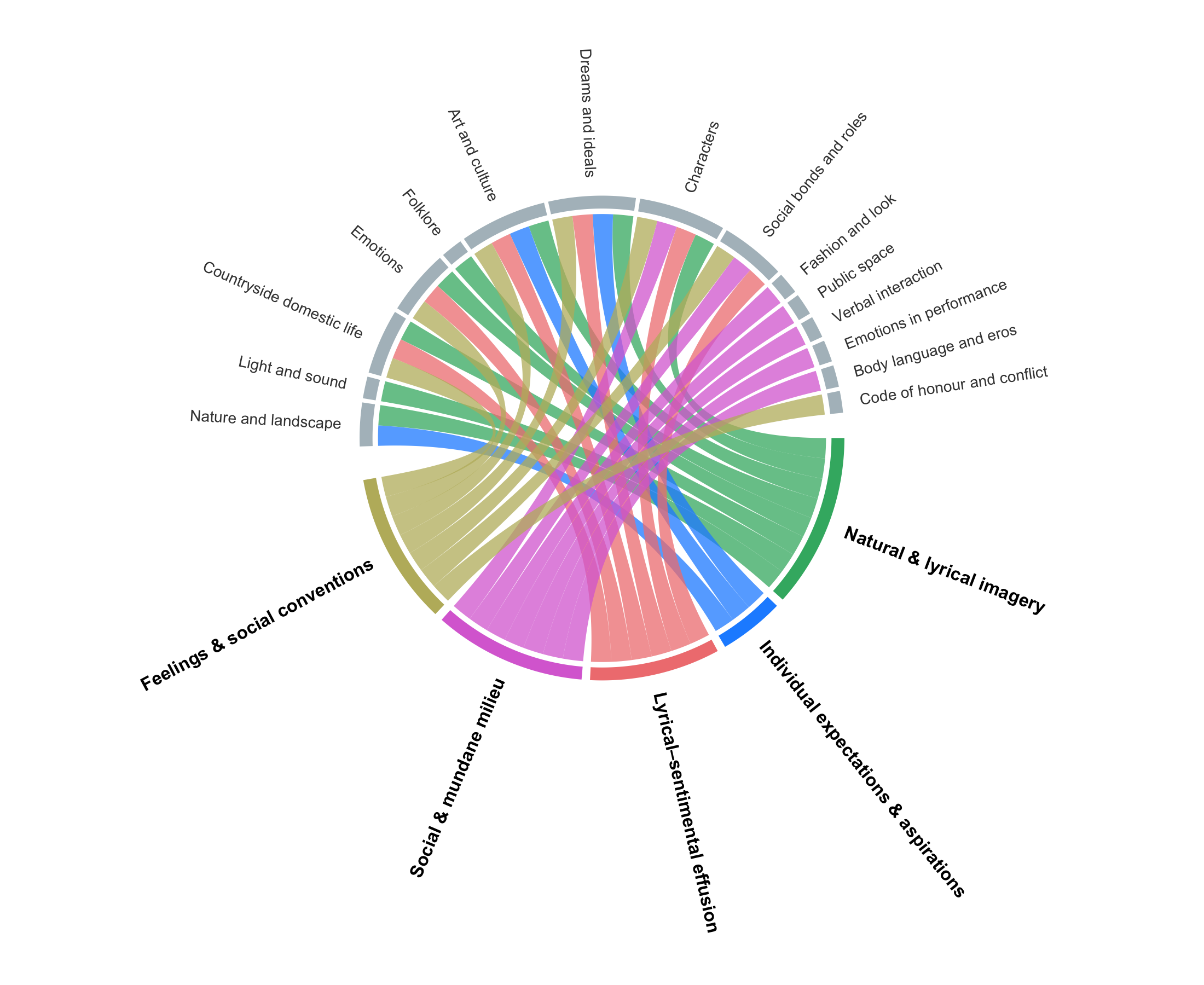}
\caption{Chord diagram showing interrelations between the five main themes (bottom) and their associated subthemes (top).}
\label{fig:chord}
\end{figure}

Finally, Fig.~\ref{fig:chord} visualises the network of interrelations between the five main themes and their associated subthemes as a circular chord diagram.

\subsection*{Narrative hubs}
\label{sec:results-hubs}

The fine-grained structure captured by the model is exemplified by contrasting the two epistolary scenes of the verse novel together with their corresponding verbal replies.
Specifically, we focus on $H_5$–$H_6$—Tat'jana’s passionate love letter to Evgenij and his subsequent reply, emphatic yet emotionally restrained—and on $H_{12}$–$H_{13}$, where the roles are reversed as Evgenij writes to Tat'jana and she answers with firmness tempered by doleful emotion (Table~\ref{tab:EO-hubs}). It is worth noting that the two epistolary episodes slightly depart from the standard stanzaic structure. Tat'jana’s letter comprises seventy-nine iambic tetrameters in free rhyme, whereas Evgenij’s letter consists of sixty iambic tetrameters with alternating rhyme. As each formally occupies a single stanza, the blocks in which they are embedded contain proportionally more lemmas than average—a deviation too small to affect model stability or interpretability.

Tables~\ref{tab:hub-Tatjana}–\ref{tab:hub-Evgenij-reply} present the hub cards for Tat'jana’s letter and Evgenij’s reply; Tables~\ref{tab:hub-Evgenij}–\ref{tab:hub-Tatjana-reply} report those for Evgenij’s letter and Tat'jana’s reply. A full example of a hub-card report is provided in Supplementary Material~B (Table~S1).
\newpage

\begin{longtable}{@{}p{0.45\textwidth}p{0.45\textwidth}@{}}
\caption{\emph{Tat'jana’s letter} — lexical hub card. sPLS–DA lemmas are denoted by \textsuperscript{\dag} or \textsuperscript{\ddag}; \textsuperscript{\dag}, novel lemma (not in LDA top lists); \textsuperscript{\ddag}, porous lemma (appearing among the top terms of another topic).}\\
\label{tab:hub-Tatjana}\\
\toprule
\textbf{Theme / Subtheme} & \textbf{Lemmas (IT)} \\
\midrule
\endfirsthead
\toprule
\textbf{Theme / Subtheme} & \textbf{Lemmas (IT)} \\
\midrule
\endhead
\multicolumn{2}{r}{\footnotesize Continued on next page}\\
\endfoot
\bottomrule
\endlastfoot
\multicolumn{2}{@{}l@{}}{\textbf{Lyrical–sentimental effusion}} \\
Art and culture & voce, pensiero \\
Characters & Evgenij \\
Countryside domestic life & casa, fuoco\textsuperscript{\ddag} \\
Dreams and ideals & anima, destino, giorno, mondo, sogno, speranza, tempo \\
Emotions & caro, cuore, dolce, incontro\textsuperscript{\dag} \\
Social bonds and roles & moglie\textsuperscript{\ddag} \\
\addlinespace
\multicolumn{2}{@{}l@{}}{\textbf{Natural \& lyrical imagery}} \\
Characters & Tat'jana \\
Dreams and ideals & sogno \\ 
Emotions & lettera \\
Light and sound & silenzio \\
Nature and landscape & cielo \\
Folklore & magico \\
\end{longtable}

\begin{longtable}{@{}p{0.45\textwidth}p{0.45\textwidth}@{}}
\caption{\emph{Evgenij’s reply} — lexical hub card.}\\
\label{tab:hub-Evgenij-reply}\\
\toprule
\textbf{Theme / Subtheme} & \textbf{Lemmas (IT)} \\
\midrule
\endfirsthead
\toprule
\textbf{Theme / Subtheme} & \textbf{Lemmas (IT)} \\
\midrule
\endhead
\multicolumn{2}{r}{\footnotesize Continued on next page}\\
\endfoot
\bottomrule
\endlastfoot
\multicolumn{2}{@{}l@{}}{\textbf{Lyrical–sentimental effusion}} \\
Art and culture & pensiero \\
Characters & Evgenij \\
Countryside domestic life & cielo\textsuperscript{\ddag}, fuoco\textsuperscript{\ddag}\\
Dreams and ideals & anima, destino, giorno, mondo, sogno, tempo, vero \\
Emotions & amore, caro, cuore, dolce, felice, gioia\textsuperscript{\ddag}, sentimento \\
Social bonds and roles & sposo, moglie\textsuperscript{\ddag}  \\
\addlinespace
\multicolumn{2}{@{}l@{}}{\textbf{Individual expectations \& aspirations}} \\
Dreams and ideals & anno, dolce, mondo, pieno, sogno, tempo \\
Nature and landscape & primavera \\
\end{longtable}
\newpage

\begin{longtable}{@{}p{0.45\textwidth}p{0.45\textwidth}@{}}
\caption{\emph{Evgenij’s letter} — lexical hub card.}\\
\label{tab:hub-Evgenij}\\
\toprule
\textbf{Theme / Subtheme} & \textbf{Lemmas (IT)} \\
\midrule
\endfirsthead
\toprule
\textbf{Theme / Subtheme} & \textbf{Lemmas (IT)} \\
\midrule
\endhead
\multicolumn{2}{r}{\footnotesize Continued on next page}\\
\endfoot
\bottomrule
\endlastfoot
\multicolumn{2}{@{}l@{}}{\textbf{Lyrical–sentimental effusion}} \\
Characters & Evgenij \\
Dreams and ideals & anima, destino, giorno, tempo \\
Emotions & amore, caro, cuore, dolce, gioia\textsuperscript{\ddag}, incontro\textsuperscript{\dag}, noia\textsuperscript{\dag} \\
\addlinespace
\multicolumn{2}{@{}l@{}}{\textbf{Feelings \& social conventions}} \\
Characters & Evgenij, Lenskij \\
Code of honour and conflict & sangue, segreto \\
Dreams and ideals & libertà\textsuperscript{\ddag}, ora, tempo \\
Emotions & allegro, triste \\
\end{longtable}

\begin{longtable}{p{0.45\textwidth} p{0.45\textwidth}}
\caption{\emph{Tat'jana’s reply} — lexical hub card.}
\label{tab:hub-Tatjana-reply}\\
\toprule
\textbf{Theme / Subtheme} & \textbf{Lemmas (IT)} \\
\midrule
\endfirsthead
\toprule
\textbf{Theme / Subtheme} & \textbf{Lemmas (IT)} \\
\midrule
\endhead
\midrule
\multicolumn{2}{r}{\emph{Continued on next page}} \\
\midrule
\endfoot
\bottomrule
\endlastfoot

\multicolumn{2}{l}{\textbf{Lyrical–sentimental effusion}} \\
Art and culture & libro \\
Characters & Evgenij \\
Countryside domestic life & casa, njanja \\
Dreams and ideals & destino, mondo, sogno, tempo \\
Emotions & amore, caro, cuore, amore, felice, passione\textsuperscript{\ddag}, sentimento \\
Social bonds and roles & moglie\textsuperscript{\ddag} \\
\addlinespace
\multicolumn{2}{l}{\textbf{Individual expectations \& aspirations}} \\
Dreams and ideals & alto, età\textsuperscript{\dag}, gloria, mondo, noto, ora\textsuperscript{\ddag}, primo, sogno, tempo, vita \\
\end{longtable}
\newpage

\section*{Discussion}
\label{sec:discussion}

\subsection*{Computational framework}
\label{sec:discussion-framework}

Applying LDA topic modelling to a single-verse novel corpus requires careful preprocessing decisions. Word morphology matters in poetry—especially in highly inflected languages such as Italian and Russian—because inflection generates families of forms that contribute subtly to style \citep{NavarroColorado2018}. Procedures such as stemming and lemmatisation reduce this variability but inevitably suppress morphological cues \citep{Schoech2017,NavarroColorado2018}. In this study, lemmatisation was adopted to control lexical sparsity and to focus the analysis on lemma-level distributions. Given the relatively small scale of the corpus, token-level modelling would risk inflating idiosyncratic forms and obscuring the thematic signal \citep{JockersMimno2013}. Although lemmatisation attenuates stylistic nuances that may be relevant in Puškin’s verse novel—many of which are unavoidably lost in translation—maintaining lexical consistency across the text was a priority. This choice should be understood as a modelling decision aimed at stabilising thematic inference, rather than as a claim that lemma-level representations exhaust the literary meaning of the verse novel.

Two further preprocessing decisions deserve comment. First, verbs were not included among the content words retained for modelling. Although verb forms are known to carry stylistic and narrative information \citep{Schoech2017,NavarroColorado2018}, their high frequency and functional role would have required systematic stop-word filtering. Given that the stop-word list was deliberately kept minimal in this study, excluding verbs offered a pragmatic way to reduce noise while maintaining thematic focus, a choice empirically supported by sensitivity runs. The exclusion of verbs should therefore be read as a strategy to privilege thematic interpretability over full lexical coverage, not as a judgement on the intrinsic literary relevance of verbal forms.

Second, proper nouns were retained, although it is common in literary topic modelling to exclude them to prevent topic domination by character names. Puškin’s verse novel makes distinctive use of proper names—not only for the main characters (\emph{Evgenij}, \emph{Tat’jana}, \emph{Lenskij}, \emph{Ol’ga}, and, to a lesser degree, \emph{Zareckij}, \emph{Triquet}, \emph{Budjanov}) but also for figures from literature and the arts (e.g., \emph{Byron}, \emph{Rousseau}, \emph{Richardson}). With the exception of the central protagonists, the frequency of such names is generally low. Moreover, consolidating the diminutives and surname/forename variants of the two main pairs—Tat’jana–Evgenij and Ol’ga–Lenskij—into a single lemma was seen to ensure consistent aggregation of character mentions without leading to excessive topic domination. This decision was guided by the expectation that major characters could act as thematic anchors or interpretive keys. After lemmatisation and entity normalisation, the vocabulary was restricted to content words (nouns, including proper nouns, and adjectives) with a minimal frequency threshold to reduce noise while retaining rare but potentially informative lemmas.

With entity normalisation in place, the next design choice concerned how to segment the verse novel into documents for LDA—a critical step in shaping the modelling framework \citep{Schoech2017}. By aggregating stanzas into blocks of ten–eleven stanzas (documents), the analysis balanced the need for local resolution against the statistical requirements of topic modelling. Too fine a segmentation would have produced unstable distributions and fragmentary themes; conversely, too coarse a segmentation would have blurred local narrative turns. Although common methodological guidelines suggest document lengths of at least 300 words per document \citep{Murakami2017}, or even up to 1{,}000 words \citep{JockersMimno2013}, these figures are often reported without specifying whether counts are before or after preprocessing. In this study, our documents contained a median of $165$ content-word tokens; this count proved sufficient to yield a robust co-occurrence signal while keeping within-text dynamics visible. Critically, larger blocks would dilute the episode-level variation necessary for analysis, whereas smaller ones (e.g., halving the block size) would reduce per-document counts and increase sparsity.

A central modelling decision concerned the choice of topic number. The Gibbs sampler used to fit the LDA model exhibited sensitivity to initial seeds and to chain length; in practice, this was mitigated by multiple initialisations, generous burn-in, and inspection of between-run stability. The label-switching problem was addressed via post-hoc alignment of topics across runs. Under this regime, and with the two hyperparameters set to $\alpha = 2$ (document–topic prior) and $\beta = 0.15$ (topic–term prior), the five-topic solution ($K=5$) proved stable across repeated runs and, crucially, yielded interpretable groupings that could be meaningfully aligned with the verse novel’s narrative (Table~\ref{tab:topics}). Alternative $K$ produced either lower stability (e.g., $K=8$) or reduced narrative interpretability (e.g., $K=3$). Hyperparameters were chosen for a small, single-work corpus rather than for large-corpus defaults. With short documents, $\alpha=2$ (Dirichlet concentration $K\alpha=10$) encourages mixed, non-spiky $\gamma$, while $\beta=0.15$ provides additional smoothing over a reduced vocabulary so that rare or idiosyncratic lemmas do not dominate topics. This departs deliberately from settings commonly used for large datasets (e.g., $\alpha=50/K$, $\beta\approx0.01$), and aligns with the stability/interpretability criteria adopted here.

The dataset produced by the segmentation strategy adopted ($35\times756$) falls in the typical small-sample, high-dimensional regime ($n \ll p$), with many more lexical features ($p$) than documents ($n$). In such settings, sPLS–DA is known to be well suited, as it performs simultaneous dimension reduction and feature selection under explicit supervision. Similar corpus-to-feature ratios have been reported in literary applications—for instance, \citet{SaccentiTenori2012} analysed the \textit{Divina Commedia}, which used $100\times371$ matrices to discriminate among \emph{cantiche} or between binary classes. In this context, the decision to parameterise the one-versus-rest (OvR) binary classifiers and the final multiclass model without cross-validation was deliberate: our aim was diagnostic validation rather than optimal prediction. Cross-validation was reserved for internal stability checks (see Table~\ref{tab:splsda-class}), whereas any reliable estimation of performance metrics—and consequently of model tuning—would have been statistically fragile given the limited sample size. The final multiclass sPLS–DA served primarily as a lexicon-consolidation step. Selecting top terms separately from the five OvR classifiers would have produced highly overlapping or asymmetric lists and hindered the definition of a common lexical core. The integrated multiclass solution, by contrast, yielded a compact and thematically balanced set of discriminant lemmas that complement the unsupervised topics.

sPLS–DA was used as a diagnostic to test whether the LDA-derived labels are recoverable under supervision. Cross-validation showed high separability of the dominant topics (Table~\ref{tab:splsda-class}), suggesting that the latent LDA partitions reflect robust lexical contrasts rather than mere sampling noise. This was particularly evident in the fairly high balanced accuracy across all topics, effectively mitigating the concern of uneven class sizes. Furthermore, the sparsity constraint of sPLS–DA yielded compact, theme-specific loadings—short lists of discriminant lemmas that offer the potential to sharpen the lexical core without altering the underlying unsupervised topic structure. The clear diagonal agreement observed in the cross-method heatmap (Fig.~\ref{fig:cross-over}) confirms the convergence between latent co-occurrence patterns and explicit supervised discrimination. The resulting core-exclusive lemmas (Supplementary Material~A) serve both as robust validation markers and concise interpretative keys, and will be revisited in the narrative mapping section.

At the same time, the present framework does not aim to capture all dimensions of literary meaning. Stylistic irony, satirical commentary, and subtleties of narrative voice often operate beyond the lexical co-occurrence patterns accessible to bag-of-words models. The objective here is more modest: to test whether a relatively simple and interpretable probabilistic framework can recover stable thematic structures that remain useful for computational close reading. Although there is scope for methodological refinement—both in the definition of the top-term lists and in the operationalisation of core-exclusive terms—the present implementation was considered sufficient for the purposes of this study. The sPLS–DA stage provided a meaningful extension of the lexical dictionaries, adding 53 discriminant lemmas to the 150 derived from the five LDA topics (203 in total). Not all additional lemmas are equally expressive from a thematic standpoint, yet the overall agreement between methods and the substantive lexical enrichment obtained were deemed satisfactory for our analytical goals.

\subsection*{LDA–topics and themes}
\label{sec:discussion-themes}

Having established the computational framework, we now turn to the thematic structure that emerged from the model and to its alignment with the verse novel’s narrative development. The five topics recovered by LDA were interpreted as coherent themes, as summarised in Table~\ref{tab:topics}. Each theme is defined by a distinctive lexical core, illustrated in the bilingual treemap of Fig.~\ref{fig:treemap}. Their longitudinal dynamics are visualised in the narrative map of Fig.~\ref{fig:narrative-map}, on which the key narrative hubs identified in Table~\ref{tab:EO-hubs} are superimposed.

\paragraph{Natural \& lyrical imagery}
Defined by lemmas such as \emph{Tat'jana}, \emph{inverno} ‘winter’, \emph{neve} ‘snow’, \emph{bianco} ‘white’, and \emph{bosco} ‘forest’, this theme conveys the natural and seasonal atmosphere through which the verse novel’s lyrical and elegiac tone comes to the foreground, positioning \emph{Tat'jana} as the central figure. Its lexical field is shaped by elemental and often solitary images—\emph{inverno}, \emph{bosco}, as well as \emph{suono} ‘sound’ and \emph{silenzio} ‘silence’—closely tied to Tat'jana’s inner world (\emph{lettera} ‘letter’, \emph{romanzo} ‘novel’, \emph{sogno} ‘dream’) and to her links with Russian folklore (\emph{luna} ‘moon’, \emph{magico} ‘magic’, \emph{orso} ‘bear’). The presence of \emph{Mosca} ‘Moscow’, \emph{slitta} ‘sleigh’, and \emph{strada} ‘road’ anchors the theme in her departure from the countryside toward urban life. Core-exclusive lemmas from sPLS–DA—\emph{lacrima} ‘tear’, \emph{pianto} ‘weeping’, \emph{grido} ‘cry’, and \emph{lume} ‘lamp’—reinforce the lyrical register and mark her emotional presence throughout the verse novel. 

This theme predominates in Chapter~5, which opens with the description of the Russian winter and Tat'jana’s dream, and re-emerges through much of Chapter~7, during her solitary wandering in the fields, her visit to Evgenij’s deserted estate, and the subsequent journey to Moscow. It is thus natural to regard \emph{Natural \& lyrical imagery} as essentially Tat'jana’s theme, whose seasonal and emotional imagery—from her love of the Russian winter and the moonlit countryside to her farewell to nature—corresponds closely to the portrait outlined in \citet{Pushkin2021}.

\paragraph{Individual expectations \& aspirations}
Defined by lemmas such as \emph{gloria} ‘glory’, \emph{mente} ‘mind’, \emph{sogno} ‘dream’, \emph{tempo} ‘time’, and \emph{vita} ‘life’, this theme gathers the vocabulary of personal projection and Romantic self-reflection. It intertwines meditations on existence (\emph{mente}, \emph{tempo} ‘time’, \emph{vita}) with the pursuit of intellectual and moral achievement (\emph{gloria}, \emph{sogno}), while maintaining a strong connection to poetry through \emph{Musa} ‘Muse’, \emph{canto} ‘song’, and \emph{verso} ‘verse’. Its markers often carry a hopeful, regenerative connotation—an aspiration to simplicity and harmony with nature (\emph{dolce} ‘sweet’, \emph{fiore} ‘flower’, \emph{fiume} ‘river’, \emph{primavera} ‘spring’). The theme thus captures a recurrent impulse toward authenticity and renewal. Relational terms (\emph{amico} ‘friend’, \emph{lettore} ‘reader’) situate these aspirations within a dialogic or public sphere. The sPLS–DA core-exclusive lemmas—\emph{amante} ‘lover’, \emph{elegia} ‘elegy’, and \emph{sacro} ‘sacred’—add an introspective dimension, aligning the language of aspiration with the emotions and sensibility of poetic reflection.

In the longitudinal map, this theme shows distinct peaks in the second half of Chapter~1, when Evgenij is portrayed as undergoing a profound existential crisis and the narrator, now his declared friend, addresses the reader directly—oscillating between nostalgic recollection of worldly life, aspiration to poetry, and rejection of Romantic clichés in favour of a simple, natural mode of existence.

\paragraph{Lyrical–sentimental effusion} 
Defined by lemmas such as \emph{amore} ‘love’, \emph{caro} ‘dear’, \emph{cuore} ‘heart’, and \emph{sogno} ‘dream’, this theme concentrates the lexicon of emotion and intimacy—feelings voiced, pleaded, or confessed—making it the most affectively charged of the five. Markers such as \emph{anima} ‘soul’, \emph{destino} ‘destiny’, \emph{pensiero} ‘thought’, \emph{speranza} ‘hope’, and \emph{tempo} ‘time’ connect its sentimental register to a transcendent dimension. The ambivalence of emotions is signalled by lexical pairs such as \emph{felice} ‘happy’ and \emph{tristezza} ‘sadness’, \emph{dolce} ‘sweet’ and \emph{lacrima} ‘tear’; the spectrum of feelings is also anchored in a literary landscape through \emph{libro} ‘book’, \emph{romanzo} ‘novel’, and \emph{verso} ‘verse’. 

The theme evokes vividly the world of Tat'jana—an association made explicit by \emph{njanja} ‘nanny’, her trusted confidante, by the reference to \emph{libro} and \emph{romanzo} ‘novel’, her instruments of sentimental education, and by domestic terms such as \emph{casa} ‘home’ and \emph{sposo} ‘groom’. The sPLS–DA core-exclusive lemmas connect especially to Evgenij—\emph{gioia} ‘joy’, \emph{noia} ‘boredom’, and \emph{passione} ‘passion’—and capture the full emotional temperature, from ecstatic joy to world-weary ennui. Interestingly, the lemma \emph{incontro} ‘meeting’ appears in this theme, serving as a lexical beacon for the moments where the two protagonists’ emotional trajectories intersect. 

In the longitudinal map, this theme dominates a substantial portion of the verse novel. It becomes especially active in Chapter~3, where Tat'jana meets Evgenij and suddenly falls in love with him. The theme reaches its two major peaks in the epistolary arcs: Tat'jana’s letter and Evgenij’s reply in Chapters~3–4, and Evgenij’s letter and Tat'jana’s reply in Chapter~8, where the emotional roles are reversed and Evgenij becomes the one who falls in love.

\paragraph{Social \& mundane milieu}
Defined by lemmas such as \emph{aspetto} ‘appearance’, \emph{ballo} ‘ball’, \emph{chiasso} ‘noise’, \emph{dama} ‘lady’, and \emph{folla} ‘crowd’, this theme revolves around social performance, spectacle, and convention. It centres on collective scenes—balls, theatres, and salons—where behaviour is ritualised and appearances dictate interaction. Lexical cues such as \emph{inchino} ‘bow’ and \emph{moda} ‘fashion’ evoke the formal codes of the social world in which Evgenij moves, while terms like \emph{discorso} ‘speech’ and \emph{parola} ‘word’ mark the prevalence of conversation and superficial exchange over genuine communication. The register also includes terms associated with affect and desire—\emph{geloso} ‘jealous’, \emph{passione} ‘passion’, and \emph{sguardo} ‘gaze’—and discreet signs of eros such as \emph{fantasia} ‘fantasy’, \emph{piedino} ‘little foot’, and \emph{senso} ‘sense’. The sPLS–DA core-exclusive lemmas provide strong anchors for this social mise-en-scène, including \emph{foggia} ‘style’, \emph{lingua} ‘tongue’, and explicit references to theatre and dance—\emph{lorgnette} ‘opera glasses’, \emph{mazurka} ‘mazurka’, and \emph{palco} ‘theatre box’. 

In the longitudinal map, this theme rises prominently in the portrayal of the dandy’s life in Saint Petersburg (Chapter~1), reappears at Tat'jana’s name-day celebration at the Larins’ home (end of Chapter~5), and returns in the Moscow and Saint Petersburg scenes of Chapters~7–8, framing the verse novel’s exploration of public life and social masquerade.

\paragraph{Feelings \& social conventions}
Defined by contrasting lemmas such as \emph{amico} ‘friend’, \emph{nemico} ‘enemy’, \emph{allegro} ‘cheerful’, \emph{triste} ‘sad’, \emph{freddo} ‘cold’, and \emph{fuoco} ‘fire’, this theme gathers the vocabulary of conflict, restraint, and confrontation. It is marked by powerful symbols of violence and honour—\emph{pistola} ‘pistol’, \emph{sangue} ‘blood’, and the sPLS–DA core-exclusive lemmas \emph{duello} ‘duel’, \emph{morte} ‘death’, and \emph{morto} ‘dead’. Other discriminant terms—\emph{alba} ‘dawn’, \emph{buio} ‘darkness’, and \emph{libertà} ‘freedom’—introduce a chiaroscuro of emotion, balancing fatalism and release. The theme foregrounds the inevitable collision between private feeling and the social codes of honour and propriety. 

The presence of three main characters—Evgenij, Lenskij, and Ol’ga—situates it within the chain of events leading to the fatal duel. The lexicon is framed by domestic and rural markers such as \emph{campagna} ‘countryside’ and \emph{casa} ‘house’, which heighten the contrast between the peaceful environment and the mounting tension. The appearance of \emph{Zareckij} among the lemmas further underscores his narrative role as a catalyst—the deus ex machina ensuring that the duel takes place without any attempt at reconciliation. In this perspective, Lenskij emerges as a passive instrument in Zareckij’s hands, while Evgenij’s indifference to social norms seals the tragic outcome. 

In the longitudinal map, this theme peaks in Chapter~6, precisely when events accelerate inexorably toward the duel between the two friends and Lenskij’s death, following the incident at Tat'jana’s name-day celebration. The biographical echo of Puškin’s own death in a duel lends retrospective poignancy to this episode of the verse novel.

\subsection*{Examples of narrative hubs}
\label{sec:discussion-examples-hub}

The four narrative hubs considered are $H_5$–$H_6$ and $H_{12}$–$H_{13}$—the two epistolary episodes ($H_5$–$H_{12}$) and their corresponding replies ($H_6$–$H_{13}$), in which Tat’jana and Evgenij exchange roles (Table~\ref{tab:EO-hubs}). It is worth noting that these epistolary episodes slightly depart from the standard stanzaic density. Tat’jana’s letter comprises seventy-nine iambic tetrameters in free rhyme, while Evgenij’s letter consists of sixty iambic tetrameters with alternating rhyme. As each formally occupies a single stanza, the corresponding blocks contain proportionally more lemmas than average, a deviation that does not affect model stability or interpretability.

Tat’jana’s letter (Table~\ref{tab:hub-Tatjana}) is dominated by \emph{Lyrical–sentimental effusion}, characterised by the subthemes "Emotions" (\emph{caro} 'dear', \emph{cuore} 'heart', \emph{dolce} 'sweet') and "Dreams and ideals" (\emph{anima} 'soul', \emph{destino} 'fate', \emph{sogno} 'dream', \emph{speranza} 'hope', \emph{tempo} 'time'). The runner-up theme, \emph{Natural \& lyrical imagery}, anchored by \emph{Tat’jana} herself within "Characters", frames the confession in her solitary and romantic world—\emph{lettera} 'letter' ("Emotions"), \emph{silenzio} 'silence' ("Light and sound"), \emph{sogno} 'dream' ("Dreams and ideals"). The sPLS–DA core-exclusive lemma \emph{incontro} 'meeting' ("Emotions") reinforces the reading of the scene as an intensely lyrical and unreserved outpouring. 

Evgenij's reply (Table~\ref{tab:hub-Evgenij-reply}) still revolves around \emph{Lyrical–sentimental effusion}, and the lexical overlap with Tat'jana’s letter is remarkable, especially across "Emotions" and "Dreams and ideals" (e.g., \emph{destino} 'fate', \emph{tempo} 'time'). Lemmas associated with \emph{Individual expectations \& aspirations} further reframe this field through a more introspective lens, marking the self-conscious control that defines Evgenij's tone. They still fall within "Dreams and ideals", yet the same subthemes acquire different tonalities when occurring under different themes: what in Tat'jana’s letter conveyed longing and emotional exposure becomes, in Evgenij's reply, a register of distance and restraint.

Evgenij's letter (Table~\ref{tab:hub-Evgenij}) still centres on \emph{Lyrical–sentimental effusion}, encompassing a full range of feelings—from \emph{amore} 'love' to \emph{gioia} 'joy' and \emph{noia} 'boredom' ("Emotions")—and the temporal marker (\emph{tempo} 'time'). A critical shift is, however, evident: the runner-up theme \emph{Feelings \& social conventions} intrudes visibly. Lexical cues related to the honour code (\emph{sangue} 'blood', \emph{segreto} 'secret' in "Code of honour and conflict") and the presence of the lemma \emph{Lenskij} in "Characters" darken the affective register, as further shown by the contrasting pair \emph{allegro} 'cheerful' and \emph{triste} 'sad' in "Emotions"—the temporal marker (\emph{tempo} 'time') is also present here. The sPLS–DA core exclusives \emph{noia} 'boredom', \emph{gioia} 'joy', \emph{libertà} 'freedom', and \emph{tempo} 'time'—the latter two belonging to "Dreams and ideals"—sharpen this ambivalence, reflecting Evgenij's struggle between emotion and restraint. Remarkably, the sPLS–DA core-exclusive lemma \emph{incontro} 'meeting' in "Emotions" recurs here as in Tat'jana’s letter, suggesting a lingering echo of emotional reciprocity. 

In Tat'jana's final reply (Table~\ref{tab:hub-Tatjana-reply}), \emph{Lyrical–sentimental effusion} once again dominates, with leading subthemes "Emotions" and "Dreams and ideals". Echoes of sentimental education—\emph{libro} 'book' in "Art and culture"—and nostalgic memories of the past—\emph{njanja} 'nanny' in "Countryside domestic life"—are also present. Many lemmas from the preceding hubs reappear (in particular those connected to the temporal dimension), yet the tone shifts markedly: Tat'jana's language moves away from \emph{Natural \& lyrical imagery}—her earlier signature theme—towards \emph{Individual expectations \& aspirations}. This shift is underscored by the influx of lemmas from "Dreams and ideals", which articulate the complex interplay of intimacy and distance defining her final, socially codified stance—\emph{moglie} 'wife' in "Social bonds and roles".

Across the four hubs, several recurrent lemmas delineate a network highly revealing of Puškin’s poetics: destino (‘fate’) and tempo (‘time’), which resonate through all four episodes, and sogno (‘dream’), absent only from Evgenij’s letter. Another fil rouge is provided by the triad amore, caro, cuore (‘love, dear, heart’). The temporal dimension is central to \emph{Evgenij Onegin}. Puškin explicitly insists on the presence of an internal calendar embedded within the poem (his note 17 to the text)—a structural device that, inter alia, can be used to anchor the flow of events to the cyclical rhythm of the seasons. Within the novel’s minimalist plot, the alternation of winter and spring—each described with lyrical intensity—marks the time and suggests the possibility of a renewal that remains, however, fundamentally denied to humankind. Or, if allowed, not without existential cost. 

This dilemma is embodied in the evolution and fate of the two protagonists \citep{Pushkin1996}. Tat’jana—so intimately attuned to the magic of nature around her—evolves from a solitary rural adolescent, nourished by sentimental novels, into a sophisticated lady of high society. Yet the price of this transformation is an excruciating experience of love, culminating in the renunciation of her dreams of affection and happiness for a life confined within the golden cage of social convention. Conversely, Evgenij—a young man incapable of feeling at home anywhere, even in the countryside—reveals a profound inability to mature beyond his youthful dandyism, remaining trapped in an existential paralysis from which he cannot break free. In a sense, Evgenij never arrives in time to express his own emotions and thus remains incapable of true renewal. As Tat’jana sadly observes in her reply to him, \emph{Then happiness seemed possible, so near! ... But now my destiny is carved in stone, immutable.}—a thought the Poet himself echoes: \emph{Alas it’s sad to think that youth was granted to us all in vain...} (Chapter~8,~stanzas~X–XI).

\section*{Conclusion}
\label{sec:conclusion}

This study showed that a hybrid topic modelling approach—combining Latent Dirichlet Allocation (LDA) with sparse Partial Least Squares Discriminant Analysis (sPLS–DA)—can yield stable and interpretable results even within the constraints of a single-verse novel corpus. The five-topic solution effectively partitioned the lexicon of Ghini’s Italian translation of \emph{Evgenij Onegin} into a set of coherent themes.

Methodologically, the study provides a validated workflow for computational close reading of highly structured literary texts, where supervised discrimination refines and confirms unsupervised inference. In the case of a single, well-known work such as \emph{Evgenij Onegin}, the aim is not large-scale discovery but the formalisation of thematic structures that attentive readers already perceive intuitively. By making these structures explicit and traceable across the poem, topic modelling provides a complementary perspective that can guide and refine close reading rather than replace it.

Future work may extend this framework to other long-form verse narratives, exploring how computational modelling can further support interpretive reading of complex poetic structures.

\end{document}